\documentclass[pdflatex,sn-mathphys-num, iicol]{sn-jnl}

\usepackage[utf8]{inputenc}
\usepackage[english]{babel}

\usepackage{amsmath,amssymb,amsfonts}
\usepackage{mathtools}
\usepackage{mathrsfs}
\usepackage{mathptmx}
\usepackage{cases}

\usepackage{graphicx}
\usepackage{graphics}
\usepackage{epsfig}
\usepackage{booktabs}
\usepackage{multirow}
\usepackage{xcolor}
\usepackage{arydshln}
\usepackage[font=small]{caption}
\usepackage{subcaption}

\usepackage{algorithm}
\usepackage{algorithmicx}
\usepackage{algpseudocode}
\usepackage{listings}
\usepackage{verbatim}
\usepackage{comment}

\usepackage[title]{appendix}
\usepackage{manyfoot}
\usepackage{lipsum}
\usepackage{setspace}
\usepackage{atbegshi}
\usepackage{textcomp}
\usepackage[normalem]{ulem}

\usepackage[nolist,nohyperlinks]{acronym}
\usepackage{enumitem}
\newlist{Aenumerate}{enumerate}{1}
\setlist[Aenumerate]{label=A\arabic*)}

\newtheorem{teorema*}{Theorem}

\hypersetup{colorlinks=true,linkcolor=blue,urlcolor=blue}

\definecolor{GraziaColor}{rgb}{0.2, 0.4, 0.6}

\begin{acronym}
    \acro{VSA}{Variable Stiffness Actuators}
    \acro{LLMs}{Large Language Models}
    \acro{VR}{Virtual Reality}
    \acro{SoE}{Sense of Embodiment}
    \acro{HRI}{Human-Robot Interaction}
\end{acronym}
\theoremstyle{thmstyleone}%

\theoremstyle{thmstyletwo}%
\theoremstyle{thmstylethree}%
\begin{document}
\title[Article Title]{\centering
Alter-Art:\\
Exploring Embodied Artistic Creation through a Robot Avatar
}


\author*[1,2]{\fnm{Do Won} \sur{Park}}\email{do.park@iit.it}

\author[2,3]{\fnm{Samuele} \sur{Bordini}}\email{samuele.bordini@phd.unipi.it}

\author[1,2,3]{\fnm{Giorgio} \sur{Grioli}}\email{giorgio.grioli@unipi.it}

\author[1,2]{\fnm{Manuel G.} \sur{Catalano}}\email{manuel.catalano@iit.it}

\author[1,2,3]{\fnm{Antonio} \sur{Bicchi}}\email{antonio.bicchi@iit.it}

\affil[1]{\orgdiv{Soft Robotics for Human Cooperation and Rehabilitation}, \orgname{Istituto Italiano di Tecnologia}, \orgaddress{\street{via Morego, 30}, \city{Genoa}, \postcode{16163}, \state{Liguria}, \country{Italy}}}

\affil[2]{\orgdiv{Centro di Ricerca ``E.Piaggio"}, \orgname{Universita' di Pisa}, \orgaddress{\street{Largo Lucio Lazzarino, 1}, \city{Pisa}, \postcode{56122}, \state{Tuscany}, \country{Italy}}}

\affil[3]{\orgdiv{Dipartimento di Ingegneria dell’Informazione}, \orgname{Universita' di Pisa}, \orgaddress{\street{Via Girolamo Caruso, 16}, \city{Pisa}, \postcode{56122}, \state{Tuscany}, \country{Italy}}}


\abstract{As with every emerging technology, new tools in the hands of artists reshape the nature of artwork creation. Current frameworks for robotics in arts deploy the robot as an autonomous creator or a collaborator, thus leaving a certain gap between the human artist and the machine. Now, we stand at the dawn of an era where artists can escape physical limitations and reshape their creative identity by inhabiting an alternative body. This new paradigm allows artists not only to command a robot remotely, but also to {\it be} a robot, to see and feel through it, experiencing a new embodied reality. Unlike virtual reality, where art is created in a digital dimension, in this case art creation is still firmly grounded in the material world: clay molded by mechanical hands, paint swept across a canvas or gestures performed on a physical stage alongside human actors. Through the robot avatar Alter-Ego, we explore the Alter-Art paradigm in dance, theater, and painting; it integrates immersive teleoperation and compliant actuation to enable a first-person creative experience. Analyzing qualitative artistic feedback, we investigate how embodiment shapes creative agency, identity and interaction with the environment. Our findings suggest that artists rapidly develop a sense of presence within the robotic body. The robot's physical constraints influence the creative process, manifesting differently across artistic domains. We highlight embodiment as a central design principle, contributing to social robotics and expanding the possibilities for telepresence and accessible artistic expression.}

\maketitle

\section{Introduction}
\label{sec:introduction}
When the pigment known as ultramarine was made from ground lapis lazuli, it was only used for the Virgin Mary's robes in religious paintings. Had Jean-Baptiste Guimet not created a synthetic version in 1826, the ``French Ultramarine", we would not have seen Van Gogh's blue sky in \textit{The Starry Night} (1889). What will we see after robotics are routinely used by artists to express themselves?

As with any new technology, innovative instruments in the hands of the artist will always lead to a new form of art. However, with the convergence of robotics and artificial intelligence in art, a new paradox appears: the more capable the machine becomes, the less the artist is present in the creation. Whether the robot serves as a collaborator in a co-creative process or an autonomous creator generating art independently, the role of the artist's bodily presence in the creative process progressively decreases. The final product is incomplete without a dimension: the human hand with all its imperfections.

The research of this work aims to address the role of the artist in reclaiming the physical presence in the creative process, not by rejecting the machine but by inhabiting it. Picasso moved from realism to cubism not as loss of interest from form, but as a return to a more primitive, natural and spontaneous way of expression, one reminiscent of a child's gaze. In a similar spirit, we consider the possibility of becoming the robot to see through its eyes, act through its body and create through its hands. The robot is no longer a mere instrument; it has become an extension of the artist himself. As an alter ego, it facilitates a new kind of physical interaction with the environment, creating an entirely new stimulus while keeping human creativity at the core. In this way, the artist transfers artistic capabilities to the robot through teleoperation and embodiment, giving birth to the concept of \textit{Alter-Art}.

This paper proposes the concept of \textit{Alter-Art} and explores it through three case studies in which professional artists interacted with the semi-humanoid robot Alter-Ego \cite{lentini2019alter}, developed through a collaboration between the Soft Robotics for Human Cooperation and Rehabilitation Lab at the Istituto Italiano di Tecnologia and the Research Center ``E. Piaggio'' of the University of Pisa, in the artistic domains of dance, theater, and painting. Through formal interviews, we examined how artists experienced the process of creating through an anthropomorphic robotic body, with a focus on embodiment, creative identity and the role of the robot's physical features. The main contribution of this paper is not the ability of the robot to create art by itself, but the exploration of the robot as a medium for embodied artistic creation: the artist does not create \textit{with} the robot, but \textit{through} it, inhabiting its body as an alternative physical self. This provides a new perspective on the relationship between social robots and art, shifting the focus from external collaboration to first-person creative experience.

Sec. \ref{sec:background} reviews related work on robots in the arts, human-robot interaction in performing and visual arts, and embodiment and telepresence. Sec. \ref{sec:Alter-Ego} describes the Alter-Ego platform, focusing on the features that enable artistic embodiment. Sec. \ref{sec: Art Experiences through Alter Ego Platform} reports on the three art experiences and the artists' perspectives. Sec. \ref{sec:discussion} discusses the findings in the context of the literature. Sec. \ref{sec:conclusion} concludes the paper.

\begin{figure}[ht]
    \centering
    \includegraphics[width=1\columnwidth]{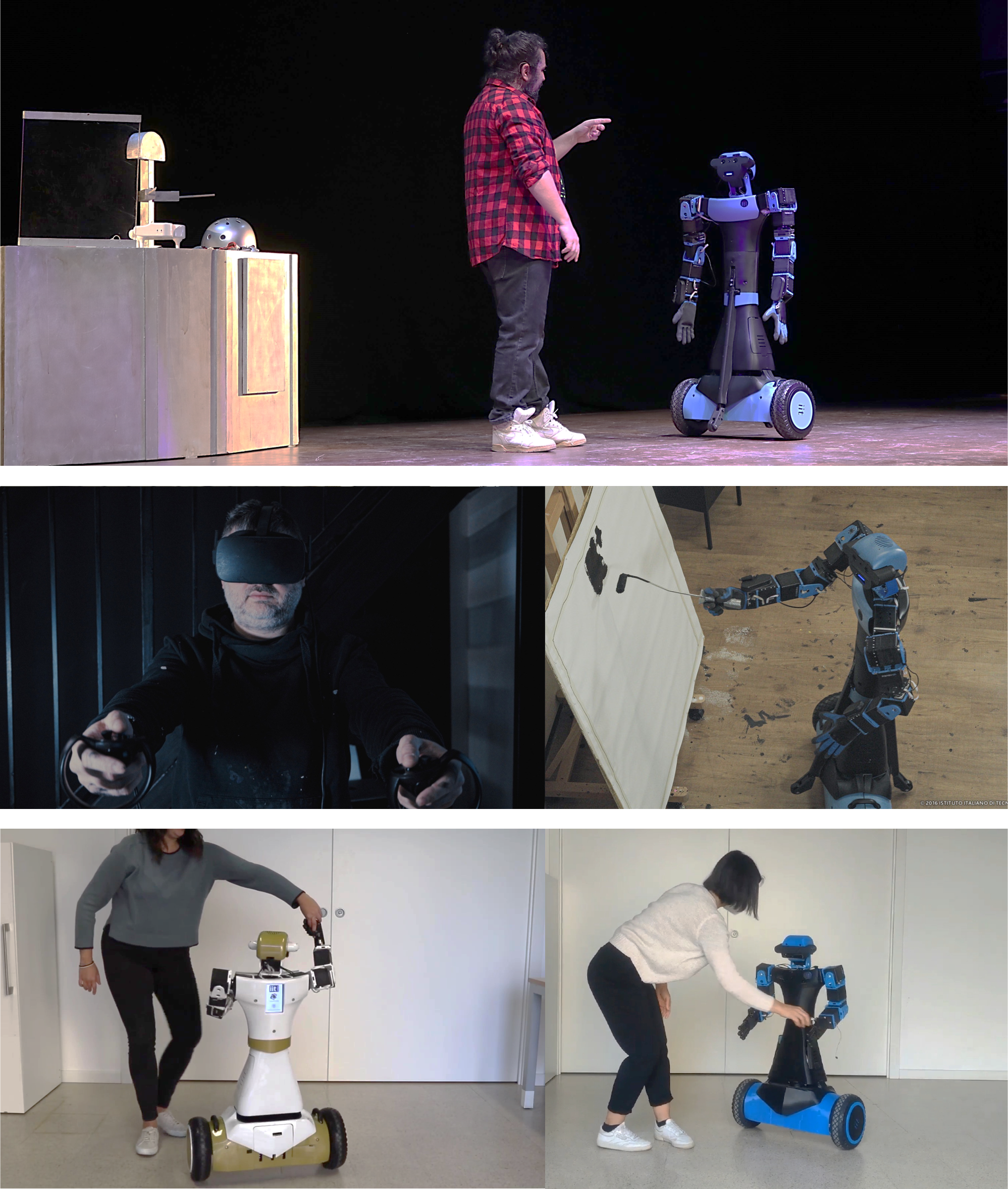}
    \captionsetup{font=footnotesize}
    \caption{Some snapshots of applications in artistic scenarios: theatre (top), painting (middle), and dance (bottom).}
    \label{fig:application}
\end{figure}

\section{Related Work}
\label{sec:background}
\paragraph{Robotics in Arts: From Mechanical Automata to Creative Agents}
\label{sec:relatedwork1}
The relationship between mechanical devices and artistic expression has ancient historical roots. Starting from Ctesibius' hydraulic organ and Hero of Alexandria's theater automata, up to Leonardo Da Vinci's \textit{Mechanical Knight} designed to entertain the court of Ludovico Sforza, machines have always proven their ability to inspire human creativity. This trend has developed into an artistic direction in the $20^{\textnormal{th}}$ century, thanks to kinetic art, promoted by Marcel Duchamp, Alexander Calder, and Jean Tinguely, and then by cybernetic art, thanks to the works of Nicolas Schöffer and Edward Ihnatowicz. An important artist in this tradition is Nam June Paik, considered the father of video art, who in 1964 created the humanoid robot \textit{Robot K-456} \cite{robotk456}, able to walk, gesture, and act, thus illustrating his idea of the combination of technology and artistic expression.

In recent decades, robotics has transformed from a part of art movements to an independent artistic style \cite{jeon2017robotic}. This transformation can be understood in terms of a taxonomy of the robot’s role in art. First, there is the robot as a theme in art, as exemplified by Sun Yuan and Peng Yu’s controversial installation \textit{Can’t Help Myself} \cite{canthelp}, in which an industrial robot constantly attempts to hold back a viscous blood-like liquid. Second, there is the robot as a tool for execution, e.g., 3D printing or robotic sculpting \cite{robotor}, where a robot realizes human input with high precision. Similarly, Tresset’s Paul series \cite{paul} explores drawing as an art form through mechanical arms that create portraits in a distinctive artistic style. Third, with the advent of artificial intelligence, there is the robot as a creator, for instance, \textit{Ai-Da} \cite{aida}, an ultra-realistic AI humanoid robot artist, and \textit{DALL-E} \cite{openai_dalle3}, a generative AI producing original artworks with minimal human input. 

However, in all of these roles, the artist delegates a creative process to the machine, maintaining only their involvement in conceiving the idea or providing the initial input. It should be pointed out that the artist is not present inside the robot but stands outside as far as the creation itself takes place through the actions of the machine. A different trajectory is identified when the artist and the robot are directly involved in the creation process. The following subsection will examine the direction of co-creation in the arts, where the artist is actively involved alongside the machine, though still physically separate from it.

\paragraph{Human-Robot Interaction in Performing and Visual Arts}
\label{sec:relatedwork2}

An emerging field of research investigates the participation of robots within artistic domains, including both robots as agents and robots as interactive partners, \textit{co-creator}, in which human and machine engage in a collaborative artistic process. For instance, interactive drawing systems \cite{schaldenbrand2024cofrida, xie2025embodied}, generative art with human input \cite{cocchella2025artists}, and improvisational robot-human dance \cite{thorn2020human}. Here, the artistic process is seen as a dialogue between human and robot, with both contributing to a final product \cite{gomez2021robot, herath2022art}. This perspective has motivated an emerging body of \ac{HRI} research examining how robots participate in specific artistic domains.
 
In \textit{dance}, Peng et al. \cite{peng2015robotic} provided a taxonomy of robotic dance within the domain of social robotics, highlighting its potential to express emotions and enhance social interaction. Thorn et al. \cite{thorn2020human} carried out research on human-robot improvised dance, revealing the possibility of co-creative movement between human and robot via mutual adaptation. Saviano et al. \cite{saviano2024multi} developed methods to map dancer aesthetic movements onto a robot arm, while Granados et al. \cite{granados2016guiding} explored physical human-robot interaction for
dance teaching. More generally, researchers such as Cuan \cite{cuan2021output} and Ladenheim et al. \cite{ladenheim2020live} have examined dance as a medium for exploring human movement, mimesis, and the boundaries between human and robot movement. Gemeinboeck \cite{gemeinboeck2021aesthetics} developed a relational-performative framework, termed \textit{bodying-thinging}, which reframes robot design through movement-based perspectives.
 
In \textit{theater}, robots have been utilized as stage performers  \cite{sovhyra2021robotic} and the concept of robot theater has been analyzed through comparative studies of human and mechanized creative processes. Donnarumma \cite{donnarumma2017beyond} introduced the concept of \textit{configuration} as a performative entity of human and nonhuman components to produce alternate forms of embodiments. However, the use of robots in theater tends to involve autonomous or pre-programmed robot behavior. The scenario of an actor fully embodying the robot via teleoperated control, as a form of theatrical mask, has not been sufficiently explored.
 
In the \textit{visual arts}, Schaldenbrand et al. \cite{schaldenbrand2022frida, schaldenbrand2024cofrida} developed FRIDA and CoFRIDA for human-robot co-painting, where the robot adapts its brush strokes based on artist feedback. CoFRIDA builds on the InstructPix2Pix model \cite{brooks2023instructpix2pix}, a pre-trained text-guided image editing framework, which is fine-tuned in a self-supervised manner to encode the robot's physical constraints and enable collaborative painting on a shared canvas. Xie et al. \cite{xie2025embodied} proposed an LLM-based robotic
arm drawing system, targeting human-robot interaction improvement. A study by \cite{cocchella2025artists} examined the perceptions of professional artists regarding collaboration with an autonomous painting robot, revealing via semi-structured interviews and thematic analysis that human-robot collaboration is experienced as both more reflective and self-directed and at the same time, more playful, compared to human-human collaboration. Another study by \cite{qin2025encountering} examined the human-robot artist relationship via its social, material and temporal aspects, revealing that the robot’s body and its malfunctioning trigger artistic exploration.

\paragraph{The Proposed Perspective: Embodiment in Art} 
A common thread across these domains is that the robot is studied as an entity external to the artist: whether as an autonomous agent, a collaborative partner, or a mechanical tool, the human always remains physically separate from the machine. Even in co-creation, the artist creates \textit{with} the robot, not \textit{through} it. What remains absent from the literature is the scenario in which the artist fully embodies the robot, perceiving through its sensors, moving through its actuators and creating through its body as an alternative physical self. This is not merely a question of teleoperation; it requires a shift from controlling an external system to experiencing the robot as one's own body in the creative act. It is precisely this gap that \textit{Alter-Art} addresses. The following situates this paradigm within the existing literature on embodiment, telepresence, and robotic avatars.

Essential to \textit{Alter-Art} is the idea of embodiment, which is described as \textit{the subjective experience of perceiving a body as one's own, controlling it and being within it.} Kilteni et al. \cite{kilteni2012sense} have provided a framework in this area by defining the concept of \ac{SoE}, which is composed of three sub-components: \textit{body ownership} (the feeling that a body is one's own), \textit{agency} (the feeling of controlling the body's actions) and \textit{self-location} (the feeling of being spatially situated within the body). The concept of embodiment has emerged from the rubber hand illusion\cite{botvinick1998rubber} and later extended to full-body virtual avatars. This framework has become central to understanding how humans relate to non-biological bodies.

In virtual reality, considerable research has shown that users can have a strong \ac{SoE} with virtual avatars that are significantly disparate from their physical bodies in terms of scale, appearance, and morphology \cite{kilteni2012sense}. Importantly, this is not just subjective; it has considerable behavioral, cognitive, and self-perceptual consequences. The question is whether this kind of embodiment can occur with physical robots in physical space. The complexity is increased as robots are part of a physical space and have physical constraints and outcomes.

The development of robotic avatar systems has advanced significantly in recent years, driven in part by the ANA Avatar XPRIZE competition \cite{zambella2025usability}, which challenged teams to develop systems enabling operators to sense, communicate, and act in remote environments as though physically present  \cite{hauser2025analysis}. Advanced robotic avatar systems like Alter-Ego \cite{lentini2019alter}, iCub3 \cite{dafarra2024icub3}, and NimbRo \cite{schwarz2021nimbro} incorporate immersive vision, touch sensing, and re-targeting of human body motion to offer users a comprehensive sense of telepresence. An extensive survey of various humanoid teleoperation methods is presented in the work of Darvish et al. \cite{darvish2023teleoperation}. In another work, Hagita et al. \cite{hagita2024cybernetic} extend the idea of humanoid teleoperation by describing cybernetic avatars as technologies that offer in-body monitoring and social interaction. An overview of telepresence social robotics is presented in \cite{almeida2022telepresence}. Among the key factors identified for robotic avatar systems is the sense of co-presence: the sensation of being socially present with another person through the robot.

Nevertheless, the majority of the existing work on robotic avatars has been devoted to the execution of functional tasks such as remote manipulation, navigation, social interaction, and assistive tasks \cite{darvish2023teleoperation, hauser2025analysis}. The question of what happens when the user experiences the embodiment of the robotic avatar not for the execution of any task but for the purpose of creating, i.e., painting, dancing, or acting, has been studied relatively less. This area is important because, unlike functional tasks, creative tasks pose specific requirements in the embodiment process, including not only control and perception, but also expressiveness, aesthetics, and improvisation. This work aims to fill the existing gap with the examination of the experience of embodiment of the robotic avatar for creative tasks by professional artists.

\begin{figure}[h]
    \centering
    \begin{minipage}{\linewidth}

        
        \begin{subfigure}[b]{\linewidth}
            \centering
            \includegraphics[width=\linewidth]{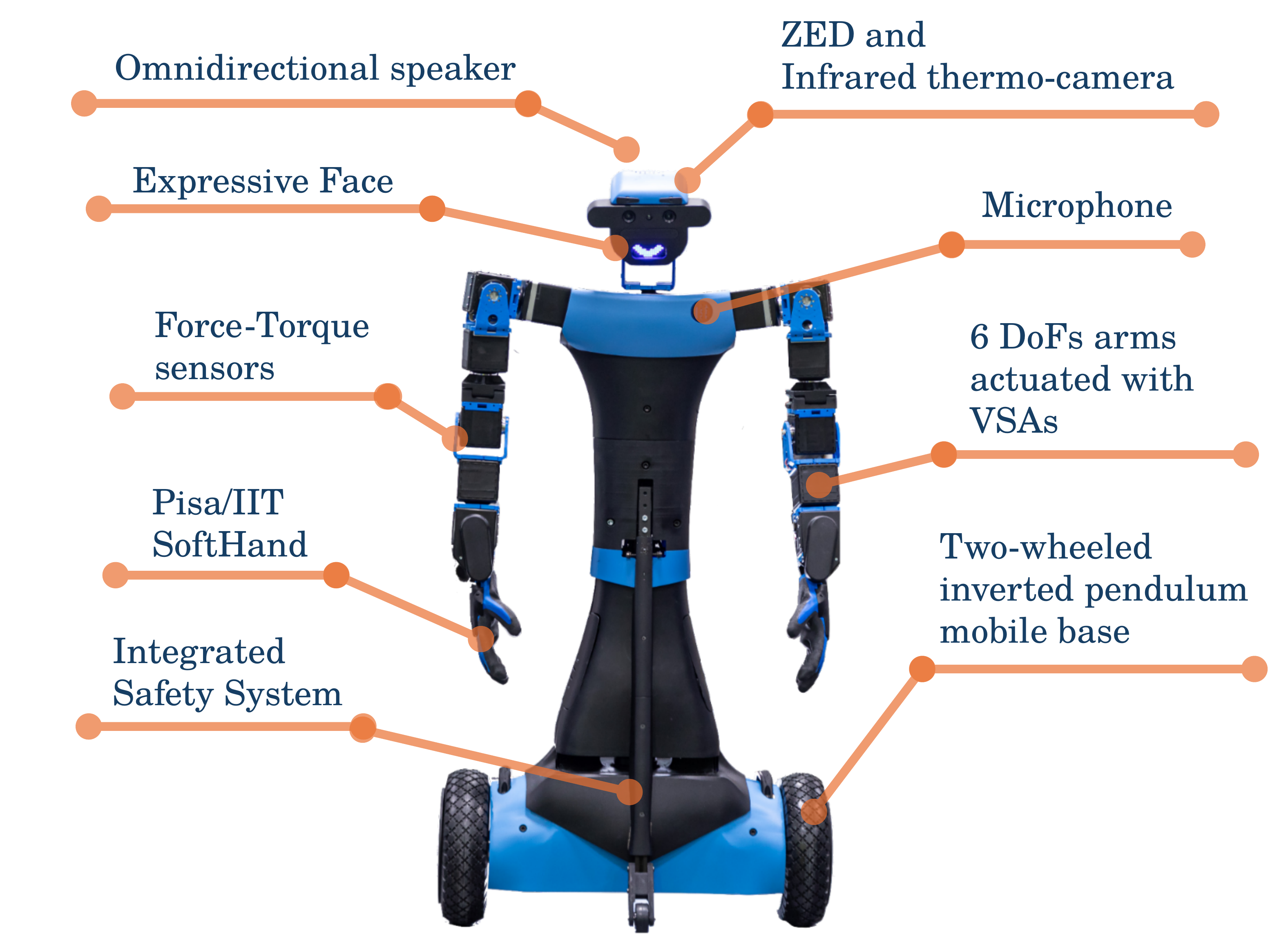}
            \caption{}
            \label{fig:alteregodesc}
        \end{subfigure}

        \vspace{0.3em}

        
        \begin{subfigure}[b]{0.475\linewidth}
            \centering
            \includegraphics[width=\linewidth]{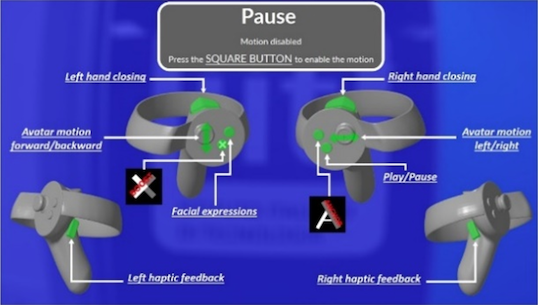}
            \caption{}
            \label{fig:alterego_b}
        \end{subfigure}\hspace{0.05cm}
        \begin{subfigure}[b]{0.4675\linewidth}
            \centering
            \includegraphics[width=\linewidth]{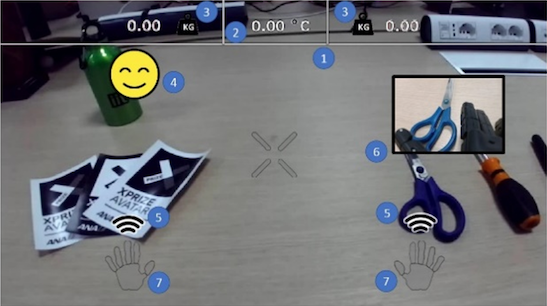}
            \caption{}
            \label{fig:alterego_c}
        \end{subfigure}

    \end{minipage}
    \captionsetup{font=footnotesize}
    \caption{The tool description: (a) illustrates the main hardware component; (b) shows the user’s view during the pause phase, (c) shows the view of the user during the teleoperation, reporting the temperature, the weight of the object the robot is manipulating, and also the emotion that the user selects.}
    \label{fig:alterego_desc}
\end{figure}

\section{The tool: Alter-Ego}
\label{sec:Alter-Ego}
\noindent Alter-Ego \cite{lentini2019alter} is a two-wheeled semi-humanoid robot. Its technical specifications are provided in Tab.~\ref{tab:alterego_specs}, and a visual description is shown in Fig.~\ref{fig:alteregodesc}. Thanks to its compliant systems and control strategy, it ensures safe and robust interaction with various environments. The platform is also accessible to users without expertise in robotics \cite{zambella2025usability}. Further details about the control interface and the user’s perspective during teleoperation are illustrated in Fig.~\ref{fig:alterego_b} and Fig.~\ref{fig:alterego_c}.

A commercial \ac{VR} headset rendered the robot's camera view, allowing the human pilot to be fully immersed in the remote environment. Alter-Ego can operate in either fully autonomous or shared autonomy modes. Most experiments were carried out under a shared autonomy framework, using a pilot station that enables a human operator to embody and control the robot remotely \cite{lentini2019alter}.

A key aspect underlies all of these features: embodiment. This condition is not solely enabled by the immersive \ac{VR} scenario, but also by several design features of the robot itself. These features, taken together, are not just engineering decisions, they are choices oriented toward enabling creativity. They determine how the artist perceives, acts and expresses through the robotic body. In the following, we describe each feature with specific attention to its role in supporting artistic embodiment. In particular, one of the main enabling factors is arm flexibility, which proves especially beneficial in artistic contexts. These actuators can behave either softly and safely or with stiffness and strength, depending on task requirements. During dance activities, it contributes to human safety, enhances the fluidity of movement, and reinforces the impression of dancing with a familiar, human-like partner.
In painting scenarios, the \ac{VSA} facilitates smooth interaction with surfaces and allows the robot to modulate its behavior by adjusting stiffness in real-time.
Likewise, in theatrical performances, the semi-anthropomorphic robot executes expressive, human-inspired gestures, contributing to a more engaging and believable stage presence.
The robot is also equipped with soft anthropomorphic hands incorporating haptic feedback, ideal for adaptive and sensitive grasping during creative tasks. While dancing, these hands ensure gentle and compliant contact with the human performer, enabling fluid and responsive physical interaction.
In painting, they enable adaptive grasping of the tools without prior knowledge of their shape or size. Notably, the hand is based on the first synergy of human hand motion, which contributes to a more natural and intuitive connection with the tools being used.

Moreover, an expressive robotic face encourages deeper empathetic engagement, particularly in theatrical and dance settings, where emotional resonance is crucial. The RGB LED matrix in the mouth, along with mobile eyebrows, helps express different emotional states such as happiness, sadness, anger, or astonishment.
\begin{table}[ht]
    \centering
    \caption{Technical description of our robotic platform.}
    \label{tab:alterego_specs}
    \renewcommand{\arraystretch}{1.2}
    \begin{tabular}{p{0.28\linewidth} p{0.65\linewidth}}
        \toprule
        \textbf{Component} & \textbf{Description} \\
        \midrule
        \textbf{Mobile Base} & Two independently actuated wheels driven by DC motors; enables smooth and fast locomotion; supports autonomous balancing. \\
        
        \textbf{Arms} & Two 6-DoF arms actuated by \ac{VSA}; capable of soft or stiff behavior, mimicking agonist–antagonist muscle dynamics. \\
        
        \textbf{Hands} & Soft anthropomorphic hands; single motor actuation with synergistic finger motion. \\
        
        \textbf{Sensors} & Surface EMG sensors to modulate arm stiffness; haptic gloves provide feedback on interaction force and texture roughness. \\
        
        \textbf{Vision System} & Dual stereoscopic camera on a neck actuated by two \ac{VSA}s (pan and tilt control); suitable for indoor and outdoor use. \\
        
        \textbf{Audio System} & Microphone in torso; speakers in head. \\
        
        \textbf{Face Expression} & $7\times16$ RGB LED matrix for mouth display and mobile eyebrows for basic emotional expression. \\
        
        \textbf{Safety System} & Integrated system to support recovery in the case of an incidental fall. \\
        
        \textbf{Pilot Station} & Lightweight wearable system: laptop, wireless router, \ac{VR} headset, haptic gloves, and joysticks. \\
        
        \textbf{Operation Modes} & Fully autonomous or shared autonomy; remote teleoperation via pilot station and \ac{VR} system. \\
        \bottomrule
    \end{tabular}
\end{table}
Finally, the artist’s point of view is streamed through commercial \ac{VR} goggles, allowing them to fully immerse themselves in the remote scene and strengthen the sense of presence within the artistic performance. When the robot is embodied, the artist sees the environment through the robot’s eyes, hears through its microphone, and perceives the sense of touch via the haptic device.
Thanks to these features, the artist can embody the robot, appropriating it as a new tool to create new forms of expression. The combination of compliant actuation, anthropomorphic hands, expressive face and immersive feedback shape a design philosophy oriented not merely toward functional teleoperation, but to support the sense of presence and creative identity that artistic embodiment requires.

\section{Art Experiences through a robotic Avatar}
\label{sec: Art Experiences through Alter Ego Platform}
\noindent 
This section shows three case studies in which professional artists engaged with Alter-Ego through dance, theater and painting, as shown in Fig.~\ref{fig:application}. The study is exploratory and qualitative in nature, aimed at capturing the artists' lived experience of creating through a robotic body. Data were collected through open conversations with the artists conducted both during and after the sessions, focusing on their perception of embodiment and the role of the robot's physical features. In the case of the painting experience with artist Melkio, additional material is gained with a short documentary filmed during the creative sessions \cite{melkiovideo}.

\subsection{Dancing experience}
\label{sec: Dancing experience}
\noindent As technology evolves, it becomes ever more accessible for artists to choreograph with machines. Dancers and performers are increasingly inviting robots onto the stage, not as props but as collaborators \cite{robotdancer}. Yet, too often, these encounters are constrained by pre-recorded motions, where the human must bend to the robot’s rhythm. But dance, at its core, is freedom, movement born of feeling, not code. In our work, we seek a more reciprocal relationship, where human and robot listen, respond, and move together in shared improvisation.

The dancer described the experience during rehearsals with these words:

\vspace{0.5mm}
\begin{quote}
\textit{``While dancing with Alter-Ego, I felt a true synergy in our movements - it wasn’t me guiding the robot or it leading me; instead, we were moving together in perfect harmony."}
\end{quote}
\vspace{0.5mm}
This feeling captures what we aimed to convey. It is not merely a performance for the audience, but the unique experience of interacting with the robot as a dance partner. Robot dancing is a kind of interactive social behavior that can be used to express emotions or intentions in a wide range of domains.

The inspiration comes from waltzer dancing, two individuals, a ``leader” and a ``follower”, dance with physical contact through their upper and lower bodies, standing in the so-called ``closed position" \cite{granados2016guiding}. The leader leads the partner and the follower has to support the leader’s intention in order to dance actively in coordination with him.

To enable this feature, we developed a novel teleoperation interface for physical human-robot interaction. The robot base, modeled as a unicycle, maps the user’s bilateral hand motions directly into linear and angular velocity commands by tracking two moving kinematic targets. Therefore, control inputs are derived from hand motion, enabling transparent robot behavior without introducing resistive dynamics or relying on explicit force sensing. The approach enforces midpoint convergence and orientation alignment, resulting in a stable and responsive motion. 
In this way, the dancer can move together with the robot, teleoperating it through their dancing intention. This enables smooth interaction and coordinated motion between the robot and the dancer. 

\begin{figure}[ht]
\includegraphics[width=0.75\columnwidth]{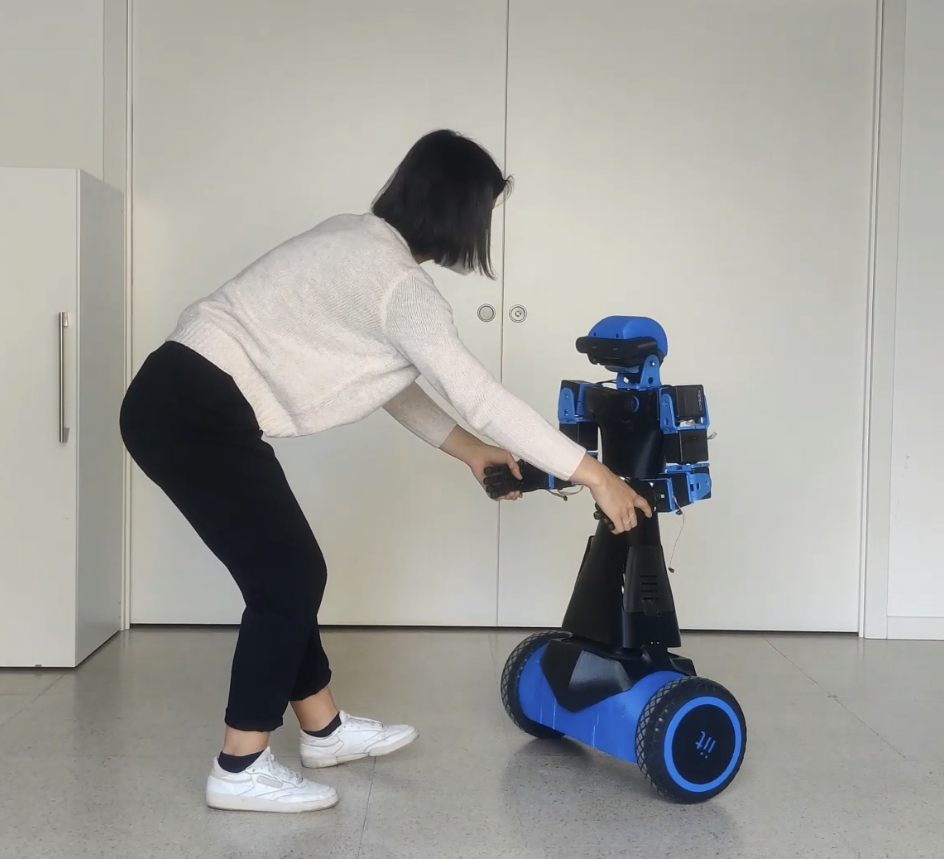}
\centering
\captionsetup{font=footnotesize}
\caption{Alter-Ego dancing together with the dancer.}
\label{fig:dancing_experience}
\end{figure}

\subsection{Theater experience}
\label{sec: Theater experience}
\begin{figure}[ht]
    \centering
    \begin{minipage}{0.95\linewidth} 

        \begin{center}
            \begin{subfigure}[b]{0.50\linewidth}
                \centering
                \includegraphics[width=\linewidth]{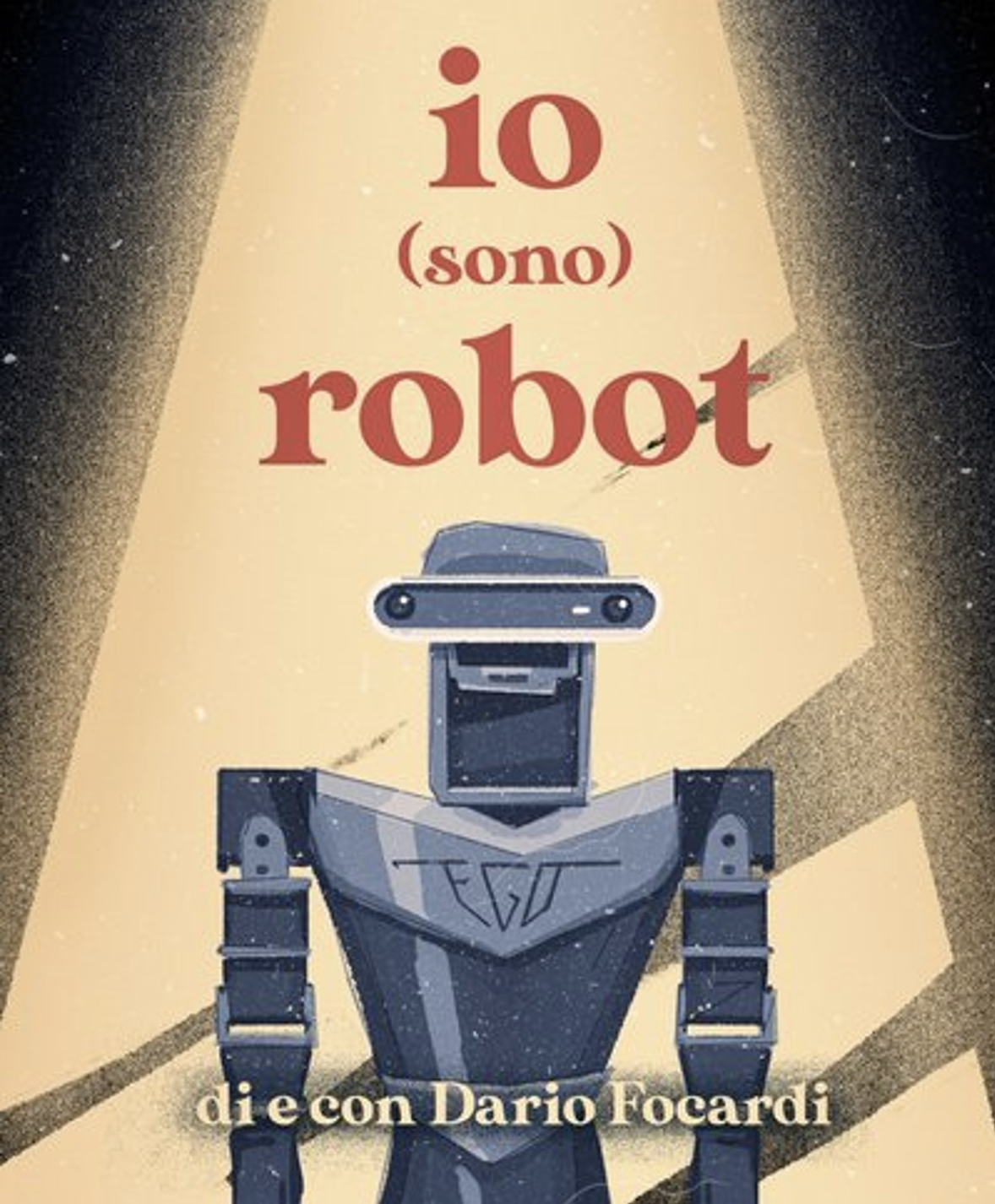}
                \caption{}
                \label{fig:teatro_a}
            \end{subfigure}
        \end{center}
        \begin{center}
            \begin{subfigure}[b]{0.52\linewidth}
                \centering
                \includegraphics[width=\linewidth]{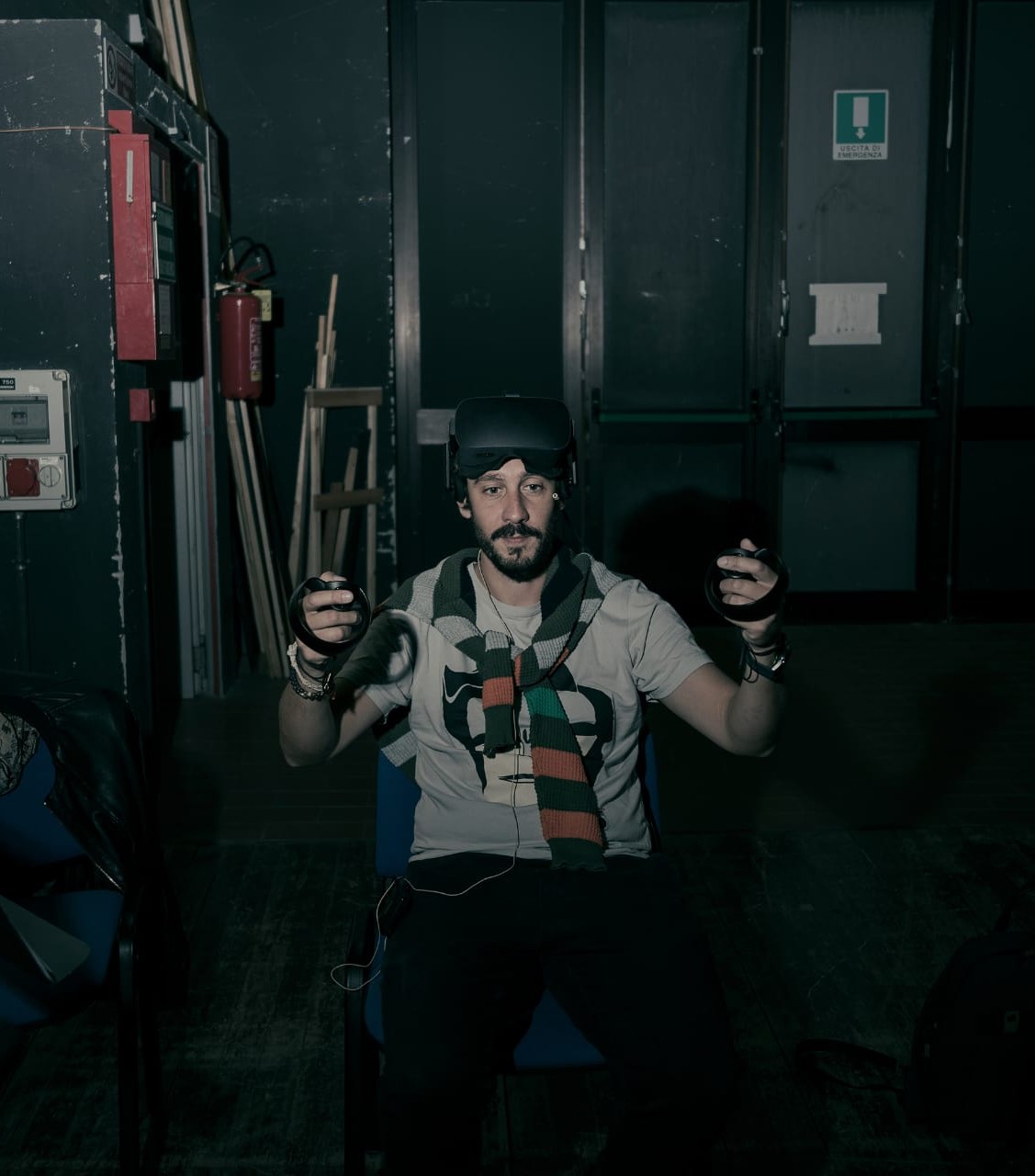}
                \caption{}
                \label{fig:teatro_c}
            \end{subfigure}\hspace{0.05cm}
            \begin{subfigure}[b]{0.424\linewidth}
                \centering
                \includegraphics[width=\linewidth]{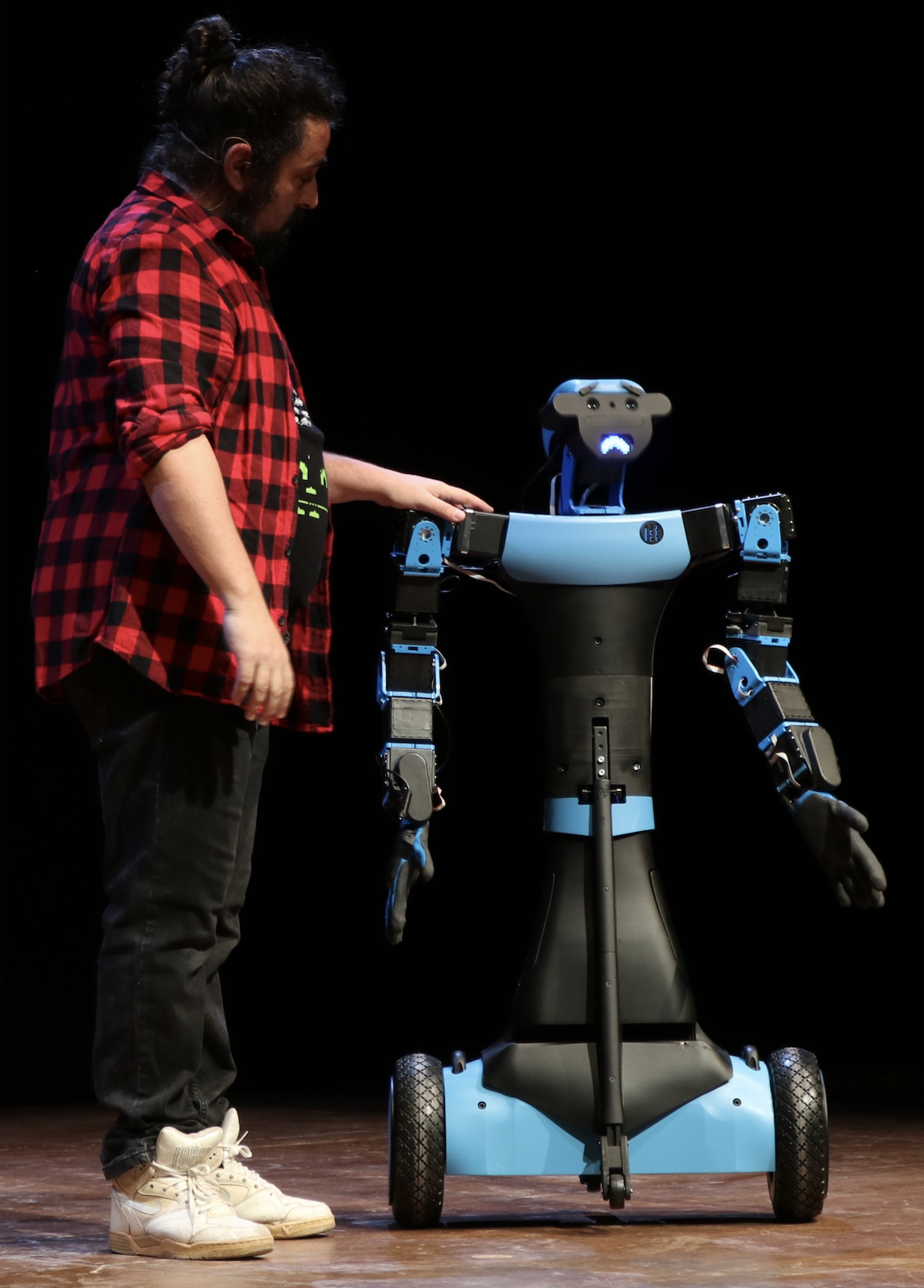}
                \caption{}
                \label{fig:teatro_b}
            \end{subfigure}
        \end{center}

    \end{minipage}
    \captionsetup{font=footnotesize}
    \caption{(a) The official poster for the theatrical performance ``Io (sono) Robot"; 
    (b) Actor Federico Raffaelli performing behind the scenes through teleoperation; (c) Robot Alter-Ego and author Dario Focardi on stage.}
    \label{fig:theater}
\end{figure}

\noindent The word \textit{robot} was born in the theatre. It was first delivered in 1920 by means of the Czech author and playwright Karel Čapek in his play R.U.R. (Rossum’s Universal Robots).
The term \textit{robot} comes from the Czech word \textit{robota}, which means forced labor or drudgery.
Čapek’s play not only gave us the term \textit{robot}, but also predicted many present-day themes, consisting of the ethics of technology and the relationship between humans and machines.
\vspace{0.5mm}
\begin{quote}
\textit{``What will be the emotional and artistic elements that will connect machines and humans? And how integrated will these two societies become?''}
\end{quote} 
\vspace{0.5mm}
These questions have inspired and continue to inspire theater director and actor Dario Focardi in his artistic and educational journey through theater.
Alter-Ego was the protagonist of the theatrical performance \textit{Io (sono) Robot}, Fig.~\ref{fig:teatro_a}, staged for the first time on September 26th, 2021, at the Città del Teatro in Pisa. The play tells the story of a human and robot friendship and features Dario Focardi on stage with the robot, the latter being animated and teleoperated by actor Federico Raffaelli. It was written for younger audiences, aged 6 to 10, as they will be the ones to interact daily with robots in the near future. 

Theater, regarded by many playwrights as a privileged medium for storytelling and self-expression, also serves as a meeting point between Art and Life. Alter-Ego’s performance is a vivid example of this. An actor can express themselves by teleoperation, using gestures, facial expressions, and speech capabilities, transforming the robot into a medium for conveying emotions and telling stories. Actor Federico Raffaelli, who animated Alter-Ego as shown in Fig.~\ref{fig:teatro_c}, described this creative process:
\vspace{0.5mm}
\begin{quote}
\textit{``Teleoperating Alter-Ego was like wearing a theatrical mask: I had to adapt my movements and emotions to a new body, exploring a unique relationship between human and robot. The voice gave soul to Ego, allowing me to convey an authentic personality to the audience, making a character visible and believable that would otherwise be impossible to portray with human presence alone. It was a constant discovery, a creative challenge that I would love to explore further.''}
\end{quote} 

\vspace{0.5mm}

From a technical perspective, the performance relied on a full-body teleoperation framework that enabled the actor to directly control the robot. In addition, the actor could teleoperate emotional states through the robot’s facial expressiveness. Specifically, Alter-Ego is capable of expressing emotions such as anger, surprise and happiness by means of an LED-based mouth and actuated eyebrows. The use of anthropomorphic hands and \ac{VSA} arms enabled safe physical interaction with other actors on stage. This compliant behavior allowed close contact and expressive physical contact without raising concerns about injuring performers or damaging the robot. Rather than wearing physical masks or costumes, the actor remotely embodied, commanding its movements and expressive features as extensions of their own, effectively transforming the robot into an embodied theatrical mask.

In Fig.~\ref{fig:teatro_b}, Dario Focardi and Alter-Ego are shown during the theater experience.

\subsection{Painting experience}
\label{sec: Painting experience}

\noindent Art often aims to reconcile technical expertise with primitive spontaneity. Pablo Picasso (1881–1973) synthesized this paradox by saying that it took him four years to paint like Raphael, but a lifetime to paint like a child. Similarly, Joan Miró (1893–1983) theorized the necessity of ``unlearning” everything he had learned in art academies in order to rebirth the purity of childlike expression. Both artists sought a form of expressiveness that only the untrained, innocent gaze of a child can offer.
Art and robotics redefine the balance between acquired skill and raw, innocent creativity. In these terms, the robot can paradoxically serve as a means to recover the primitive spontaneity that the experienced artist often struggles to preserve.

\begin{figure}[ht]
    \centering
    \begin{minipage}{0.95\linewidth} 

        
        \begin{subfigure}[b]{0.95\linewidth}
            \centering
            \includegraphics[width=\linewidth]{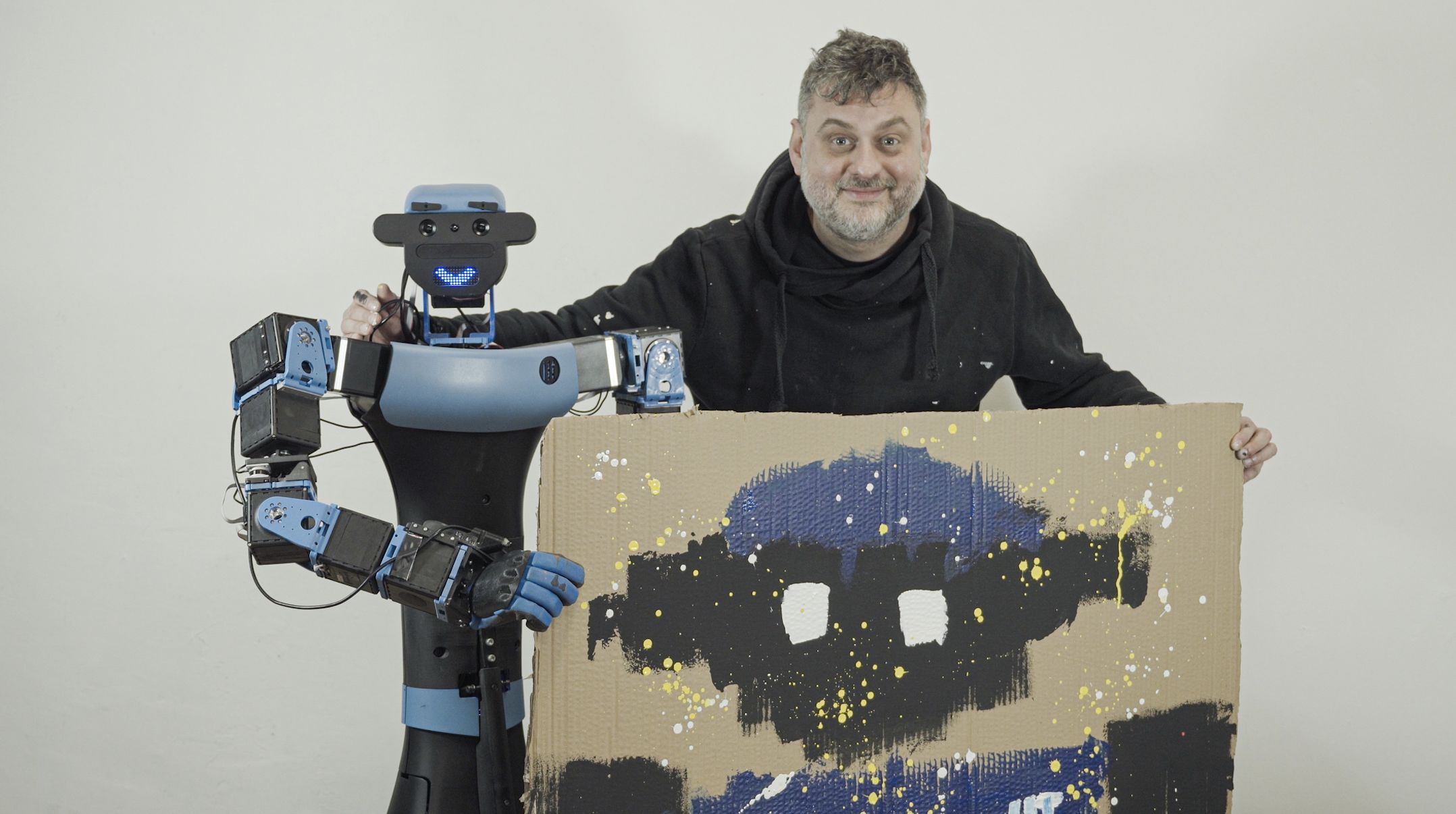}
            \caption{}
            \label{fig:melkio1}
        \end{subfigure}

        \vspace{0.3em}

        
        \begin{subfigure}[b]{0.475\linewidth}
            \centering
            \includegraphics[width=\linewidth]{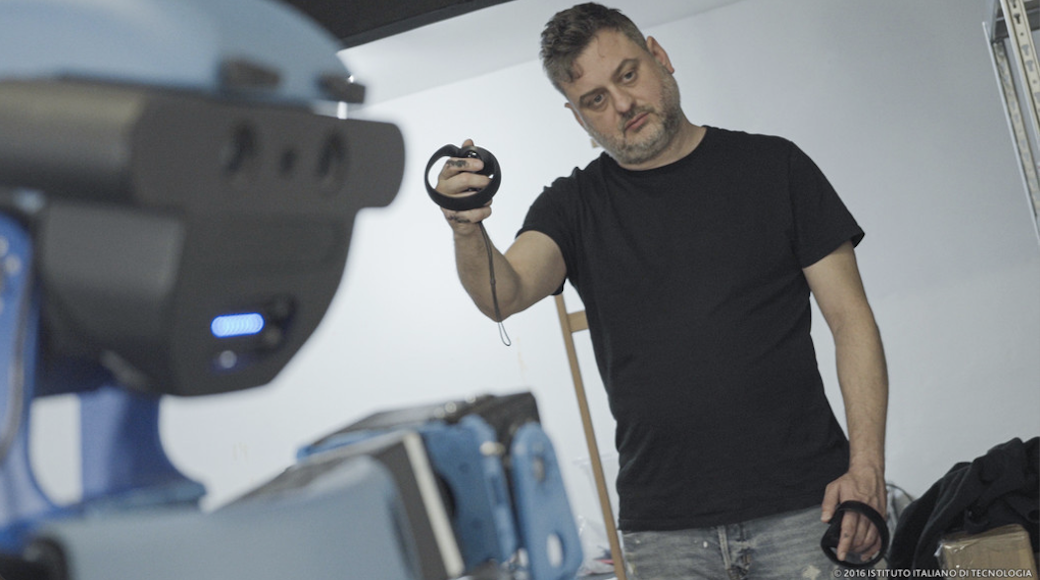}
            \caption{}
            \label{fig:melkio2}
        \end{subfigure}\hspace{0.05cm}
        \begin{subfigure}[b]{0.4675\linewidth}
            \centering
            \includegraphics[width=\linewidth]{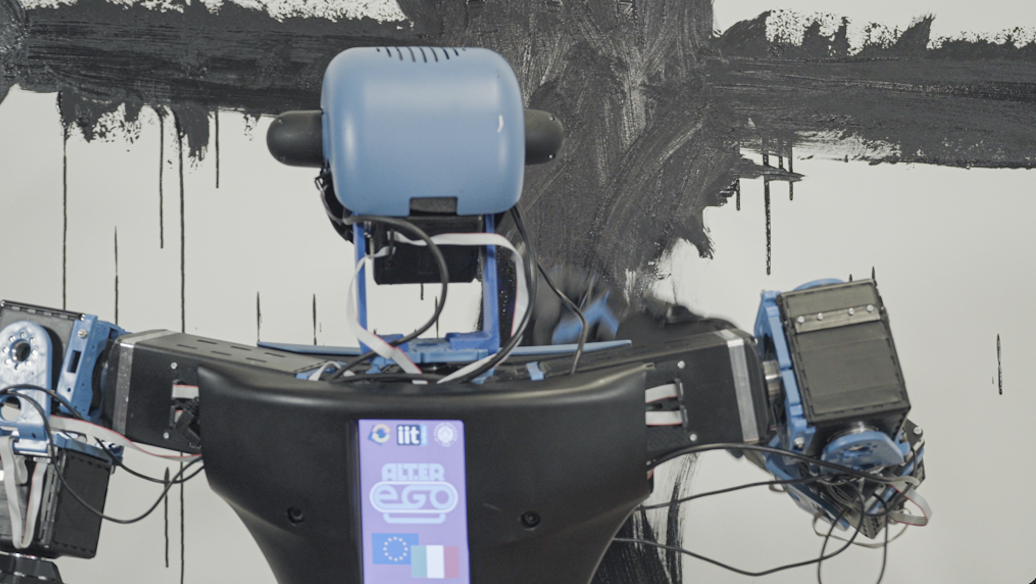}
            \caption{}
            \label{fig:melkio3}
        \end{subfigure}

    \end{minipage}
    \captionsetup{font=footnotesize}
    \caption{(a) Alter-Ego and Artist Melkio; 
    (b) Artist Melkio drawing through teleoperation; 
    (c) Robot drawing.}
    \label{fig:melkio}
\end{figure}

\begin{figure}[ht]
\includegraphics[width=0.95\columnwidth]{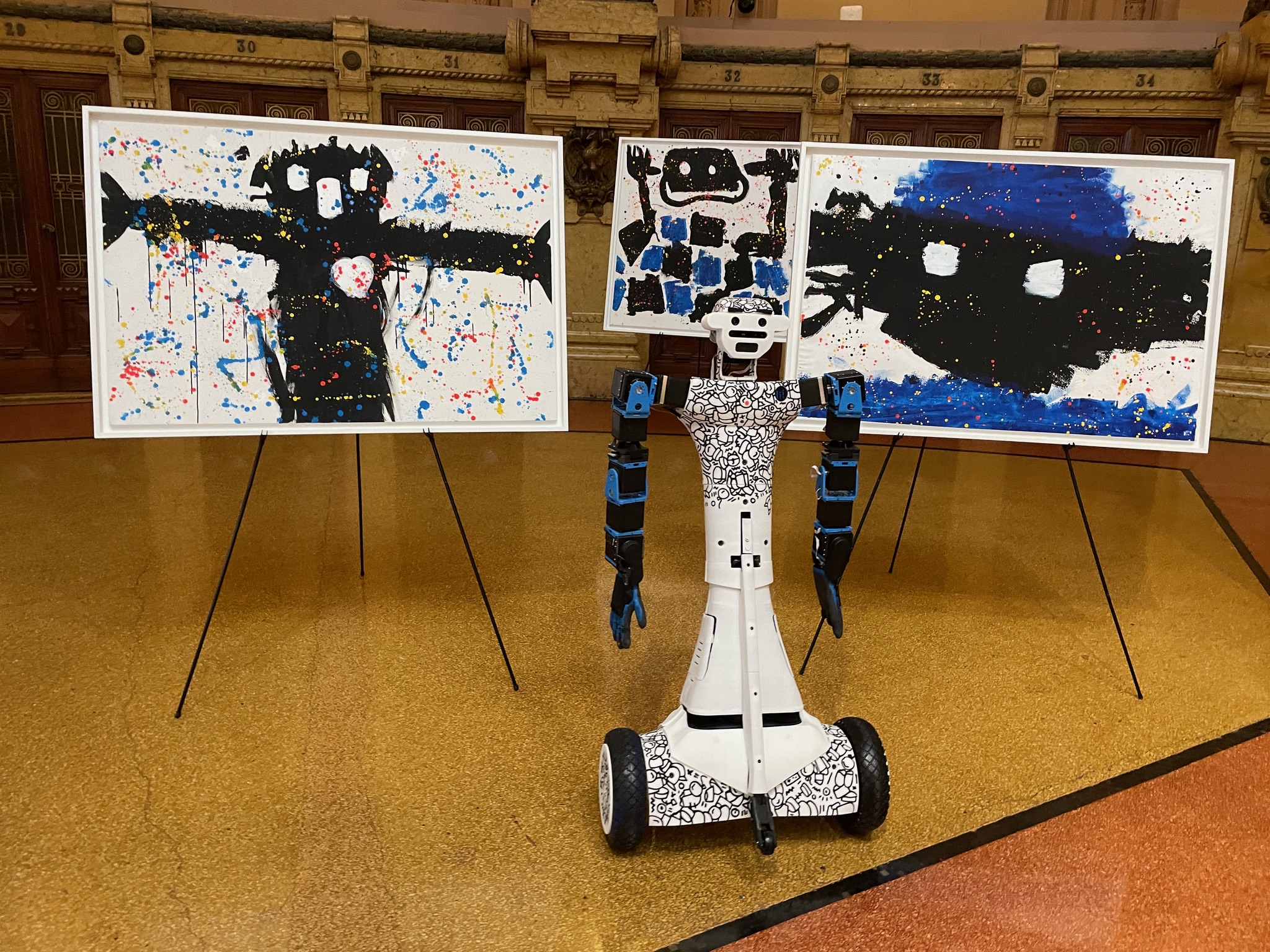}
\centering
\captionsetup{font=footnotesize}
\caption{The Human + Robot Painting project (Melkio + Alter-Ego) presentation on February 16, 2024, at Palazzo della Borsa in Genoa.}
\label{fig:painting_experience1}
\end{figure}

The Human + Robot Painting project (Melkio + Alter-Ego) Fig.~\ref{fig:painting_experience1}, was presented on February 16, 2024, at Palazzo della Borsa in Genoa, and on October 12 at Musaless Art Hotel in Verona, Melkio Art Studio, and the Camera di Commercio of Genoa. The artist Melkio has the opportunity to paint eight different artworks on paper with Alter-Ego, all featuring the characteristic scribbled drawings typical of children:

\vspace{0.5mm}
\begin{quote}
        \textit{``After the first encounters with the Alter-Ego robot, I realized the extraordinary opportunity to paint with something that is, in some way, opposite to what is expected. Alter-Ego, although a complex machine, allowed you to paint in a very instinctive way. It was as if I had the opportunity to paint through a 3-year-old. Therefore, I decided to represent simple things, seen with a simple gaze, like that of a child."}
\end{quote} 
\vspace{0.5mm}
With these words, the artist portrayed the robot as an innovative artistic tool that can help add new expressive capabilities. The robot introduces an \textit{``instinctive"} component, a characteristic strongly associated with humans and typically not linked to robotic components. 

Melkio fully teleoperated the robot during the performances (Fig.~\ref{fig:melkio2}), while preserving its expressive autonomy.
The system combines an intuitive pilot station and sensory substitution with the compliant arm and hand \cite{catalano2014adaptive}. Autonomous stabilization of the mobile platform supports a human-centred artistic project.
During painting, the mobile platform has to autonomously regulate its distance from the canvas based on the region addressed by the artist’s hand. This ensured continuous and unconstrained motion over the surface.
In parallel, the stiffness of the arm actuators enabled an intrinsic impedance in the interaction with the canvas.
Fig.~\ref{fig:melkio} shows Melkio and the Alter-Ego during the painting experience.

\section{Discussion}
\label{sec:discussion}
\begin{quote}
    \textit{``I perceive the use of robotics in the art world as one of the many tools an artist can use to express themselves. The more tools available, the more creativity can develop in entirely new ways, leading to the creation of new expressive forms and languages."}
\end{quote} 

\vspace{0.5mm}

With these words, the artist Melkio captured the core vision that stands behind \textit{Alter-Art} project: the robot, much like the choice between oil and acrylic paints, becomes a tool that shapes the creative process in its own distinctive way.
 
One of the common factors that were found in all three experiences is the immediate familiarity and immersion that each artist developed with the robot. After gaining insight into the basics of how the robot works, the transition from controlling an external machine to feeling present within it occurred naturally, enabling a strong sense of spontaneity. This transition resonates with the \ac{SoE} framework by Kilteni et al. \cite{kilteni2012sense}, particularly in terms of agency and self-location. The visual feedback system and low-latency control provided a strong sense of agency whereas the use of first-person perspective contributed to self-location within the robotic body. Although body ownership could not be quantitative observed, the artists' qualitative reports suggest a functional form of identification with the robot, sufficient to support spontaneous and expressive action. 
This indicates that full biological realism may not be necessary for effective embodiment in creative contexts.
 
Each domain provided unique insights. 
In \textit{dance}, the interaction redefined the concept of a dance partner; the intuitive control interface allowed the dancer to sidestep the resistive dynamics of the technology, achieving improvised, fluid synergy of movement.
In \textit{theater}, the actor found the teleoperated body to be similar to working with a theatrical mask: a contemporary extension of a practice rather than a radical divergence. 
With the \textit{painting} domain, the artist gradually adapted his intentions to the robot's physical capabilities, incorporating its limitations into the creative process and producing artworks with a childlike spontaneity reminiscent of the ``unlearning'' sought by Picasso and Mir\'{o}.
 
These results suggest that the embodiment of a robotic avatar is shaped by the conventions of each artistic domain. However, a common thought bridges them: the artists did not see the robot as an obstacle or a mere tool but as a body with which the artist came to identify. Robot features, like compliance and morphological differences from human body, did not prevent any form of creativity. However, all of the above qualities helped to generate art in a manner that would make sense for the artists involved. This faces the belief that increased fidelity automatically implies a better user experience. From a design point of view, the principles behind Alter-Ego (arms with varying stiffness, anthropomorphic hands, expressive face, and immersive feedback) appear to support not just manipulation and control, but also the development of a creative interaction between human and machine. In other words, it can be seen that in this context the robot is not merely an interface, but a medium through which the creation of the artwork happens.

More generally, \textit{Alter-Art} contributes to the field of social robotics by proposing embodiment as a central paradigm for creative human–robot interaction. Whereas previous work has emphasized functional aspects of teleoperation, collaboration or autonomous behavior, this study places emphasis on the more experiential and expressive aspects of human-robot interaction, providing avenues for exploration within the convergence of art, robotics, and human experience.

Finally, this work is exploratory in nature. Further investigation may include structured evaluations of the embodiment and creative experiences through studies involving larger and more diverse participant groups, and investigate how different design parameters affect the experience. Another potential avenue may include extending this approach to participants with motor disabilities, thus giving them access to new avenues of artistic creativity.

\section{Conclusion}
\label{sec:conclusion}
In this paper, the concept of \textit{Alter-Art} is proposed as the use of a robotic avatar as a medium for creating art, and it is discussed by means of three case studies in dance, theater, and painting with artists using the humanoid robot Alter-Ego. We showed that the artists quickly achieved a sense of immersion and creative agency through the robotic body. Each art form revealed a different form of embodiment: free movement in dance, masked identity in theater, and adaptive spontaneity in painting. 
Across all experiences, the robot was perceived as a body that the artist identified with.

Beyond its empirical contributions, this study poses an important question for the broader field: Is creating through an alternative body a new form of art that modifies the physical process but preserves the artist's identity? These preliminary results show not only that it is indeed possible, but that the physical characteristic of the machine may serve as a basis for its shaping. As artists have always found new voices through new instruments, from synthetic pigments to digital media, robotic embodiment could be just one more chapter in this ongoing dialog between human creativity and technological innovation.

\backmatter

\bmhead{Acknowledgements}

The authors would like to warmly acknowledge the contribution of Eleonora Sguerri and Cristiano Petrocelli. \hfill

\section*{Supplementary Material}
The supplementary material includes a video and three paintings.
\texttt{AlterArt\_1.mp4} is a short video showcasing all three artistic experiences (dance, theater, and painting), offering a comprehensive overview of the robot's role across different creative domains.
The following three works were painted by Melkio using Alter-Ego as an artistic medium. \texttt{AlterArt\_2.jpeg}: \textit{Self Portrait}. The robot gazes at its own reflection, becoming aware of its forms and colors. This self-portrait captures the moment in which the machine, guided by the artist's hand, confronts its own image as if looking into a mirror.
\texttt{AlterArt\_3.jpeg}: \textit{Humanity}. Here, the robot turns its gaze outward, toward the human beings that surround it. Drawing from its lived experience among scientists, researchers, and a curious public, the work portrays a world that welcomes the machine with benevolence.
\texttt{AlterArt\_4.jpeg}: \textit{Essence}. A deeper look into the robot's inner substance. Its modular composition, stripped to its core, reveals itself as a simple arrangement of cubes. Yet, through the hands and the VR headset, this raw structure acquires not only form but identity, delineating the very foundations of the machine's being.

\section*{Declarations}

\bmhead{Funding} This research received no external funding.
\bmhead{Conflict of interest} The authors declare that they have no conflict of interest.
\bmhead{Ethics approval and consent to participate} This study did not require formal ethics board approval. All participating artists provided informed consent to participate in the artistic experiences and interviews.
\bmhead{Consent for publication} All participants provided consent for the publication of their interviews, images, and the use of their names in this manuscript.
\bmhead{Data availability} Not applicable.
\bmhead{Materials availability} The hardware designs and components required to build the Alter-Ego robot are available through the Natural Machine Motion Initiative (NMMI) at \url{https://www.naturalmachinemotioninitiative.com/platform}.
\bmhead{Code availability} The source code of the Alter-Ego robot platform is available at \url{https://github.com/NMMI/AlterEgo}.

\bibliography{biblio}

\end{document}